\documentclass[lettersize,journal]{IEEEtran}
\usepackage{amsmath,amsfonts}
\usepackage{algorithmic}
\usepackage{array}
\usepackage[caption=false,font=normalsize,labelfont=sf,textfont=sf]{subfig}
\usepackage{textcomp}
\usepackage{stfloats}
\usepackage{url}
\usepackage{verbatim}
\usepackage{graphicx}
\usepackage{xcolor}
\usepackage{cite}

\usepackage{booktabs}
\usepackage{multirow}
\usepackage{makecell}
\usepackage[dvipsnames,svgnames]{xcolor}

\usepackage[table]{xcolor}
\usepackage{colortbl}

\definecolor{bestred}{RGB}{244,198,198}
\definecolor{secondorange}{RGB}{245,220,190}
\definecolor{thirdyellow}{RGB} {242,236,180}

\def\BibTeX{{\rm B\kern-.05em{\sc i\kern-.025em b}\kern-.08em
    T\kern-.1667em\lower.7ex\hbox{E}\kern-.125emX}}
\usepackage{balance}
\begin{document}
\title{GenSplatCodec: Feed-Forward Gaussian Splatting Compression via One-Step Diffusion  }
%GenSplatCodec: Feed-Forward Gaussian Splatting Compression via Geometry-Guided One-Step Diffusion
%GenSplatCodec: End-to-End Feed-Forward Gaussian Splatting Compression with One-Step Diffusion / GenSplatCodec: Taming One-Step Diffusion for Extreme Gaussian Splatting Compression /SplatCodec-G: Generative Gaussian Splatting Compression with One-Step Diffusion / G3Codec: Extreme 3D Gaussian Compression with One-Step Generative Decoding / SplatCodec: Feed-forward 3D Gaussian Splatting Compression with One-Step Generative Decoding  /G3Codec: Feed-forward 3D Gaussian Compression with One-Step Generative Decoding
\author{Qiang~Hu, \textit{Member, IEEE},~Zhenlong~Wu,~Lei~Huang,~Zihan~Zheng, ~Xiaoyun~Zhang, \textit{Member, IEEE},~Wenjun~Zhang, \textit{Fellow, IEEE}
\thanks{This work is supported by National Natural Science Foundation of China (62571322, 62431015), STCSM (24ZR1432000, 22DZ2229005), 111 plan (BP0719010),  and State Key Laboratory of UHD Video and Audio Production and Presentation.

Qiang Hu, Zhenlong Wu, Lei Huang, Zihan Zheng,  Xiaoyun Zhang and Wenjun Zhang are with the Cooperative Medianet Innovation Center,  Shanghai Jiao Tong University,  Shanghai,  200240,  China.
}}

\markboth{Journal of \LaTeX\ Class Files,~Vol.~18, No.~9, September~2020}%
{How to Use the IEEEtran \LaTeX \ Templates}

\maketitle

\begin{abstract}

Feed-forward 3D Gaussian Splatting (3DGS) enables scalable scene reconstruction without per-scene optimization, yet produces dense Gaussians that are costly to store and transmit. Existing feed-forward Gaussian compression methods formulate decoding as deterministic representation recovery, which becomes inadequate at low bitrates when high-frequency textures and view-dependent appearance are discarded. Although generative models offer a promising alternative, using them as standalone post-processing decouples generation from the transmitted scene structure, thereby compromising cross-view consistency. To address these limitations, we propose \textbf{GenSplatCodec}, a unified feed-forward Gaussian codec that reformulates low-bitrate Gaussian compression as geometry-guided generative decoding. We present a detail-aware feed-forward Gaussian coding scheme within a dual-stream formulation, where the resulting compact Gaussian structural stream is complemented by a lightweight reference appearance stream. We further introduce a geometry-guided one-step generative decoding approach that jointly exploits decoded structural and appearance cues through hierarchical geometry control to reconstruct high-fidelity and view-consistent novel views.  Finally, we develop a three-stage optimization strategy that stabilizes the learning of the unified codec and adapts the generative decoder to codec-derived structural and appearance cues. Extensive experiments across multiple datasets demonstrate that GenSplatCodec consistently achieves superior rate-distortion (RD) performance over existing methods.

\end{abstract}

\section{Introduction}                       %Immersive multimedia applications, such as free-viewpoint video, augmented and virtual reality, telepresence, and digital twins, demand efficient 3D scene acquisition and transmission. Practical 3D multimedia systems must rapidly reconstruct high-fidelity scene content from sparse observations, transmit it with a compact bitstream, and render novel views with low latency at the receiver side. Meeting these requirements simultaneously remains challenging under limited bandwidth and computational resources.

Immersive multimedia applications, such as free-viewpoint video, augmented and virtual reality, telepresence, and digital twins, are driving the transition from conventional 2D media to interactive 3D content. These applications require photorealistic novel-view rendering while operating under stringent bandwidth, storage, and latency constraints. Consequently, compactly representing and efficiently transmitting high-fidelity 3D scenes has become a fundamental challenge for practical immersive multimedia systems.

Recent neural rendering techniques, particularly Neural Radiance Fields (NeRF)~\cite{mildenhall2021nerf} and 3D Gaussian Splatting (3DGS)~\cite{kerbl20233d}, have significantly advanced novel-view synthesis. With its explicit scene representation and real-time rendering capability, 3DGS has become an attractive representation for immersive multimedia. However, conventional neural rendering methods typically rely on costly per-scene optimization to achieve high-fidelity reconstruction, limiting their scalability to diverse scenes. Feed-forward 3D reconstruction methods alleviate this bottleneck by directly recovering scene geometry or neural scene representations from sparse input views~\cite{wang2024dust3r,wang2025vggt,yu2021pixelnerf}. More recently, feed-forward 3DGS methods~\cite{charatan2024pixelsplat,chen2024mvsplat,xu2025depthsplat} directly predict renderable Gaussian representations without test-time optimization, substantially improving reconstruction efficiency and scalability. Nevertheless, these methods generally produce a large number of Gaussian primitives with high-dimensional attributes, resulting in considerable storage and transmission overhead that limits their deployment in bandwidth-constrained multimedia scenarios.

%To reduce the storage and transmission cost of feed-forward Gaussian representations, recent methods have explored compact representations and learned compression schemes\cite{song2026tinysplat,yu2026codecsplat}.  These approaches reduce redundancy in Gaussian primitives or entropy-code intermediate latent features, improving the practicality of feed-forward 3DGS for multimedia transmission. However, they still mainly rely on deterministic reconstruction from the compressed representation. At extremely low bitrates, high-frequency textures, view-dependent appearance, and weakly observed details are easily removed from the bitstream, making deterministic Gaussian rendering insufficient. A straightforward remedy is to apply generative enhancement to the decoded renderings, but such post-processing is decoupled from the 3D coding system: the generative model does not participate in representation design, bit allocation, or structure-aware decoding. As a result, it may produce suboptimal perceptual quality and geometry-inconsistent details across viewpoints. This motivates a unified extreme 3D codec that jointly learns compact structural transmission and geometry-guided generative synthesis.

To reduce the storage and transmission cost of feed-forward Gaussian representations, recent methods have explored compact representations and learned compression schemes~\cite{song2026tinysplat,yu2026codecsplat}. These approaches reduce redundancy in Gaussian primitives or entropy-code intermediate latent features, improving the practicality of feed-forward 3DGS for multimedia transmission. However, they still mainly rely on deterministic reconstruction from the compressed representation. At sufficiently low bitrates, high-frequency textures, view-dependent appearance, and weakly observed details are easily removed from the bitstream, making deterministic Gaussian rendering insufficient. A straightforward remedy is to apply generative enhancement to the decoded renderings, but such post-processing is decoupled from the 3D coding system: the generative model does not participate in representation design, bit allocation, or structure-aware decoding, and often relies on iterative multi-step sampling. As a result, it may produce suboptimal perceptual quality and geometry-inconsistent details across viewpoints.

We therefore revisit the formulation of feed-forward Gaussian compression. Rather than requiring a compact Gaussian representation to deterministically retain every visual detail, we argue that low-bitrate coding should preserve reliable scene structure and compact appearance evidence, while recovering missing details through geometry-guided generation. This shifts feed-forward Gaussian compression from deterministic representation recovery to geometry-guided generative decoding. Under this formulation, compact Gaussian coding and generative reconstruction are unified within the codec, allowing generative priors to improve perceptual quality while remaining explicitly constrained by the transmitted scene structure. 

Building on this formulation, we propose \textbf{GenSplatCodec}, a unified feed-forward Gaussian codec for geometry-guided one-step generative decoding. Given sparse input views and camera parameters, GenSplatCodec encodes each scene into two bitstreams: a compact Gaussian structural stream and a lightweight reference appearance stream. The former preserves 3D structure, visibility, and coarse appearance, whereas the latter supplies complementary appearance cues that are expensive to retain in compact Gaussian representations. Unlike conventional 3DGS codecs that compress per-scene optimized Gaussian representations, GenSplatCodec unifies feed-forward Gaussian prediction, compact coding, and generative novel-view reconstruction without scene-specific optimization. 

We realize this formulation through three key designs. First, a detail-aware feed-forward Gaussian coding scheme combines dual-branch Gaussian prediction, rate-distortion-guided Gaussian compaction, and geometry-attribute decoupled coding to produce the compact Gaussian structural stream, while a lightweight reference appearance stream is coded in parallel. Second, a geometry-guided one-step generative decoding method integrates hierarchical geometry control and cross-view latent interaction to jointly exploit the decoded structural and appearance cues, enabling perceptually faithful and view-consistent novel-view reconstruction. Third, a three-stage optimization strategy stabilizes the optimization of the entire codec and adapts the generative decoder to codec-derived structural and appearance conditions. Compared to the state-of-the-art method YoNoSplat~\cite{ye2025yonosplat}, GenSplatCodec achieves comparable reconstruction fidelity with approximately $\boldsymbol{17\times}$ lower storage cost.

Our main contributions are summarized as follows:

\begin{itemize}
    \item  We reformulate low-bitrate feed-forward Gaussian compression as geometry-guided generative decoding and propose \textbf{GenSplatCodec}, a unified dual-stream codec comprising a compact Gaussian structural stream and a lightweight reference appearance stream.
    %We reformulate low-bitrate feed-forward Gaussian compression as geometry-guided generative decoding, and propose \textbf{GenSplatCodec}, a unified codec that transmits compact Gaussian structures together with coded reference appearance for geometry-guided one-step reconstruction.
    %We reformulate extreme feed-forward Gaussian compression from deterministic representation recovery to geometry-conditioned generative decoding, and propose \textbf{GenSplatCodec}, a unified codec that integrates compact Gaussian transmission with geometry-guided one-step generative reconstruction.
    \item We present a detail-aware feed-forward Gaussian coding scheme that combines dual-branch Gaussian prediction, rate-distortion-guided compaction, and geometry-attribute decoupled coding to produce a compact Gaussian structural stream.
    %We develop a compression-oriented feed-forward Gaussian encoder that transforms dense Gaussian predictions into a compact structural bitstream through coding-friendly parameterization, rate-distortion-aware primitive selection, and geometry-attribute-separated entropy coding.
    %We develop a rate-distortion-aware feed-forward Gaussian encoder that reduces representation redundancy through compression-oriented parameterization, rate-distortion-aware primitive selection, and geometry-attribute-separated entropy coding.
    %We develop a rate-distortion-aware feed-forward Gaussian encoder that converts dense predicted Gaussians into a compact structural bitstream. It reduces redundancy through compression-oriented parameterization and primitive selection, and further improves coding efficiency by separately encoding Gaussian geometry and attributes. 
    %We develop a rate-distortion-aware feed-forward Gaussian encoder that produces compact Gaussian representations through compression-oriented parameterization and rate-distortion-aware primitive selection, followed by geometry-attribute-separated entropy coding for efficient transmission.
    \item We introduce a geometry-guided one-step generative decoding method with hierarchical geometry control and cross-view latent interaction. We further develop a three-stage strategy for stable optimization of the entire codec.
    %We introduce a geometry-guided one-step generative decoder with hierarchical geometry control, which reconstructs perceptually faithful novel views from compressed Gaussian structures and coded reference appearance in a single forward pass.
    %We introduce a geometry-guided one-step generative decoder with hierarchical geometry control and a three-stage optimization strategy, which reconstructs perceptually faithful novel views from compressed Gaussian structures and coded reference appearance in a single forward pass.
    \item Extensive experiments on DL3DV and RealEstate10K demonstrate that GenSplatCodec achieves superior RD performance while maintaining strong perceptual quality and cross-view consistency over state-of-the-art methods.
    
    %Extensive experiments on DL3DV and RealEstate10K demonstrate that GenSplatCodec achieves favorable rate-distortion-perception trade-offs and strong multi-view consistency with substantially smaller bitstreams than the state-of-art compression methods.
\end{itemize}

\section{Related Work}

\subsection{Novel View Synthesis}
Novel view synthesis renders unseen viewpoints from captured images and is fundamental to 3D scene reconstruction and immersive media. Neural Radiance Fields (NeRF) and their variants~\cite{mildenhall2021nerf,barron2021mip,muller2022instant,zheng2024jointrf,barron2022mip,zheng2024hpc} model scenes as continuous volumetric radiance fields and achieve photorealistic rendering through differentiable volume rendering. More recently, 3D Gaussian Splatting (3DGS)~\cite{kerbl20233d} has emerged as an efficient explicit representation that models scenes with anisotropic Gaussian primitives and supports high-quality real-time rendering through differentiable rasterization. Subsequent extensions further improve rendering quality, training efficiency, geometric accuracy, and scalability to complex scenes~\cite{huang20242d,yu2024mip,lin2024vastgaussian,wu20244d,ren2024octree,guedon2024sugar}. However, both NeRF and conventional 3DGS typically rely on scene-specific optimization, limiting their applicability to scalable and low-latency 3D multimedia systems.

\subsection{Feed-forward 3D Gaussian Splatting}
%To avoid costly per-scene optimization, recent studies\cite{yu2021pixelnerf,chen2021mvsnerf} have explored feed-forward 3D reconstruction from sparse input views. Instead of optimizing a representation for each scene, these methods learn a generalizable model that directly predicts a renderable 3D representation at inference time. 

To avoid costly per-scene optimization, recent studies~\cite{yu2021pixelnerf,chen2021mvsnerf} learn generalizable models that directly predict renderable 3D representations from sparse views. In the context of 3DGS, representative methods such as Splatter Image~\cite{szymanowicz2024splatter}, pixelSplat~\cite{charatan2024pixelsplat}, and MVSplat~\cite{chen2024mvsplat} predict Gaussian primitives from one or multiple views, enabling fast reconstruction and real-time rendering without test-time optimization. Subsequent methods improve feed-forward Gaussian prediction through enhanced depth estimation, cross-view feature aggregation, pose robustness, and geometry-aware regularization~\cite{zhang2024gs,tang2024lgm,xu2025depthsplat,hong2024pf3plat,ye2025no,wang2025vggt,jiang2025anysplat,ye2025yonosplat}. These works improve reconstruction efficiency and make 3DGS more suitable for sparse-view novel view synthesis. However, their primary objective is rendering quality rather than compact transmission, and the predicted Gaussian representations are often dense and high-dimensional, leading to substantial storage and transmission overhead.

Recent works further incorporate generative priors to enhance feed-forward Gaussian rendering, restoring high-frequency details, degraded textures, and wide-baseline novel views after initial reconstruction~\cite{lu2025prosplat,lin2025diffsplat,chen2024mvsplat360,dong2026one}. However, these methods focus on reconstruction enhancement rather than 3D scene compression. Their generative modules are typically appended after Gaussian prediction or rendering, without joint optimization with bitstream formation, rate allocation, or entropy-constrained representation design. Consequently, the generated details may be inconsistent with the transmitted 3D structure, making the overall framework suboptimal for bandwidth-constrained multimedia transmission.

\subsection{3D Gaussian Compression}

The large storage footprint of optimized 3DGS has motivated extensive research on Gaussian compression. Existing methods reduce redundancy through Gaussian pruning, spherical harmonic reduction, attribute quantization, vector quantization, and compact neural representations~\cite{fan2024lightgaussian,lee2024compact,liu2025efficientgs,zheng20264dgcpro,navaneet2024compgs,niedermayr2024compressed,chen2025hac++,hu20254dgc,hu2025vrvvc}. More recent approaches further exploit spatial dependencies through context modeling and entropy coding. HAC~\cite{chen2024hac} uses a hash-grid-assisted context model to capture spatial correlations, while ContextGS~\cite{wang2024contextgs} combines anchor-level autoregressive modeling with hyperprior features for more efficient entropy coding. Although these methods achieve high compression ratios while preserving rendering quality, they typically assume that a high-quality Gaussian representation has already been obtained through per-scene optimization. They therefore do not directly address feed-forward compression, where Gaussian prediction, representation compactness, and coding efficiency must be optimized jointly.

Recent works incorporate representation design and learned coding into feed-forward 3DGS pipelines. TinySplat~\cite{song2026tinysplat} reduces geometric, spatial, and perceptual redundancy in predicted Gaussians, while SparseSplat~\cite{zhang2026sparsesplat} adapts Gaussian density to scene structure and local information content. CodecSplat~\cite{yu2026codecsplat} entropy-codes intermediate Gaussian-generation features and decodes them into depth and Gaussian parameters, avoiding direct coding of irregular output primitives. Despite improving compression efficiency, these methods rely on deterministic recovery of Gaussian attributes, latent features, or rendered views. At low bitrates, such decoding struggles to recover high-frequency textures, view-dependent appearance, and weakly observed details. In contrast, GenSplatCodec integrates entropy-constrained Gaussian coding, coded reference appearance, and geometry-guided one-step generative decoding within a unified codec, enabling compact structural coding and perceptually faithful novel-view reconstruction.

\section{Proposed Method}
\label{sec:method}

\begin{figure*}[!ht]
    \centering
    \includegraphics[width=\linewidth]{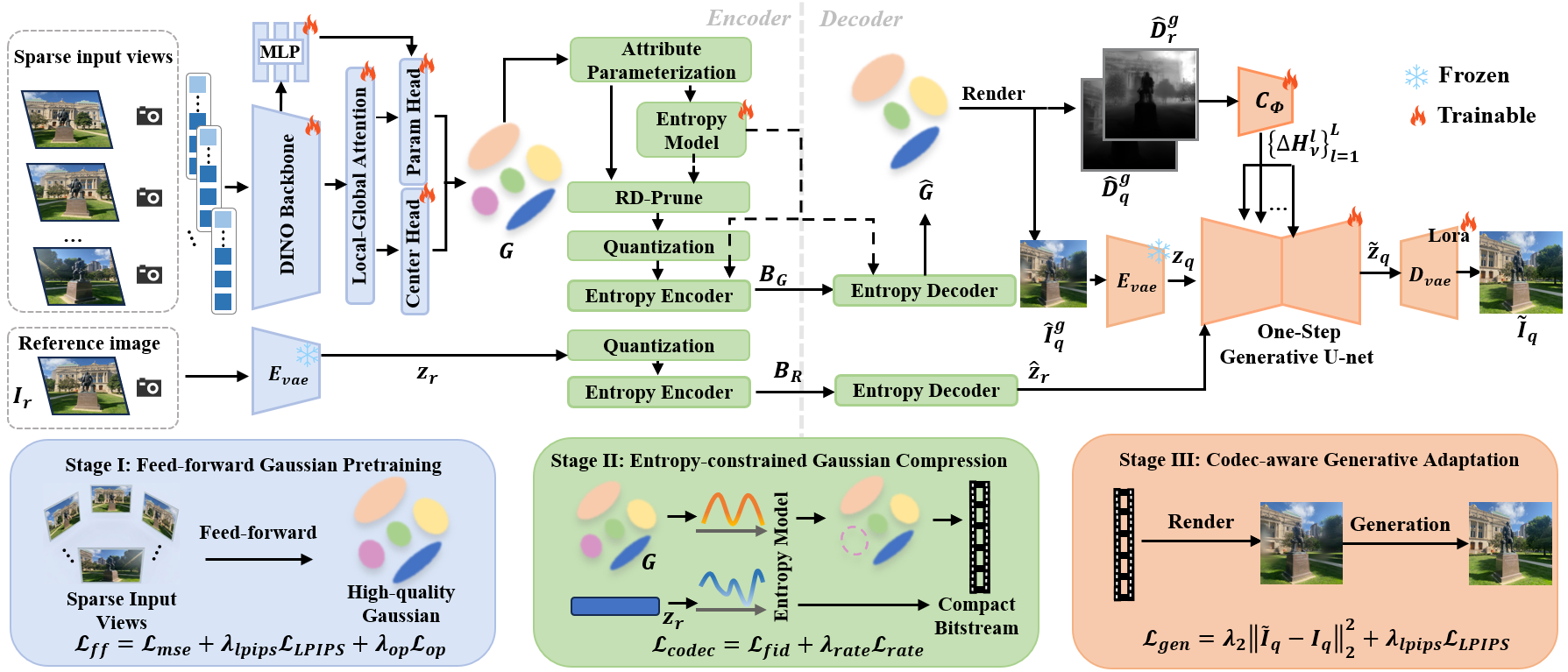}
    \caption{Overview of GenSplatCodec. Given sparse input views, the feed-forward encoder produces a Gaussian structural bitstream $B_G$ and a reference appearance bitstream $B_R$, both of which are included in the total bitrate. At the decoder, the reconstructed Gaussians provide a coarse query-view anchor and query/reference geometry cues, while the decoded reference latent supplies complementary appearance information. These codec-derived conditions are jointly processed by a geometry-guided one-step generative decoder to reconstruct the target view $\tilde{I}_q$. The bottom panel summarizes the three-stage optimization strategy, including feed-forward Gaussian pretraining, entropy-constrained Gaussian compression, and codec-aware generative adaptation.}
    %The overall framework of GenSplatCodec. Given sparse input views, the feed-forward encoder predicts and compresses a 3D Gaussian representation into the structural bitstream $\mathcal{B}_{G}$ through compression-oriented parameterization, rate-distortion-aware pruning, and geometry-attribute entropy coding. A selected reference view is encoded in the VAE latent domain to form the appearance bitstream $\mathcal{B}_{R}$. At the decoder, the reconstructed Gaussians provide a low-rate query-view anchor and geometry cues, while the decoded reference latent supplies complementary appearance information. These codec-derived conditions are jointly processed by a geometry-guided one-step generative decoder to reconstruct the final query view $\tilde{I}_{q}$.}
    \label{fig:overview}
\end{figure*}

\subsection{Overview}
\label{subsec:overview}

GenSplatCodec targets low-bitrate novel-view synthesis from sparse multi-view observations. Given a set of input views and their camera parameters,
\begin{equation}
    \mathcal{I}=\{(I_i,\pi_i)\}_{i=1}^{N},
\end{equation}
our goal is to reconstruct a high-fidelity query view $\tilde{I}_q$ at an arbitrary camera pose $\pi_q$ from compact transmitted bitstreams. The main challenge is that, under a tight rate budget, a feed-forward Gaussian representation can preserve the coarse 3D structure of the scene, but cannot explicitly retain all high-frequency textures, view-dependent appearance, and weakly observed details. To address this issue, GenSplatCodec formulates 3D scene compression as a generative Gaussian decoding problem: compact Gaussian coding preserves reliable scene structure, while codec-derived generative synthesis reconstructs perceptual details that are difficult to transmit explicitly.

As shown in Fig.~\ref{fig:overview}, GenSplatCodec adopts a dual-stream coding architecture. From the sparse input views, the encoder $\mathcal{E}$ produces two transmitted bitstreams,
\begin{equation}
(B_G,B_R)=\mathcal{E}(\mathcal{I}),
\end{equation}
where $B_G$ denotes the Gaussian structural stream and $B_R$ denotes the reference appearance stream. The Gaussian stream is generated through detail-aware feed-forward Gaussian prediction, rate-distortion-guided Gaussian compaction, and geometry-attribute decoupled coding. It carries compact 3D structure, visibility, and coarse appearance. The reference stream encodes a selected reference view in the VAE latent domain, providing complementary appearance cues that are expensive to preserve in the Gaussian domain. Both streams are transmitted as part of the codec bitstream and are included in the final bitrate.

%where $B_G$ denotes the Gaussian structural stream and $B_R$ denotes the reference appearance stream. The Gaussian stream is generated through detail-aware feed-forward Gaussian prediction, rate-distortion-guided Gaussian compaction, and geometry-attribute decoupled coding. Specifically, the retained Gaussian centers are quantized and encoded using a geometry coder, while the remaining attributes are compressed using hierarchical entropy modeling. The resulting stream preserves compact 3D structure, visibility, and coarse appearance. In parallel, the reference stream encodes a selected reference image in the VAE latent domain, providing complementary appearance cues that are expensive to retain in compact Gaussian attributes. Both streams are transmitted as part of the codec bitstream and are included in the reported rate.

At the decoder side, $B_G$ is first decoded into a compact Gaussian representation $\hat{G}$, which can be rendered at the target and reference viewpoints to provide low-rate RGB and geometry conditions. Meanwhile, $B_R$ is entropy-decoded into the reference appearance latent $\hat{\boldsymbol{z}}_r$. These decoded signals are jointly fed into a geometry-guided one-step generative decoder $\mathcal{D}$:
\begin{equation}
      \tilde{I}_q=\mathcal{D}(B_G,B_R,\pi_q,\pi_r),
\end{equation}
where $\pi_r$ is the reference camera pose. The decoder transfers reference appearance under the geometric constraints imposed by $\hat{G}$ and synthesizes the final target view in a single generative step. Since all structural and appearance conditions are reconstructed from the transmitted bitstreams, GenSplatCodec does not rely on raw reference images or unaccounted appearance side information.

%The key design principle is to decouple \emph{structural coding} from \emph{detail synthesis}. Instead of spending excessive bits to encode every appearance detail into Gaussian attributes, GenSplatCodec allocates the Gaussian stream to compact and view-consistent 3D structures, and leverages the reference stream to provide complementary appearance cues. The one-step generative decoder then recovers perceptually faithful details under explicit geometry guidance. This dual-stream generative formulation enables high-quality free-viewpoint reconstruction under tight rate constraints while maintaining a fully feed-forward pipeline without scene-specific optimization or iterative diffusion sampling.

The key idea is to separate \emph{structural coding} from \emph{detail synthesis}. The Gaussian structural stream preserves compact, view-consistent 3D structure and coarse appearance, while the lightweight reference appearance stream provides complementary appearance cues that are difficult to retain in compact Gaussian representations. Conditioned on these codec-derived structural and appearance cues, the geometry-guided one-step generative decoder reconstructs perceptually faithful details under explicit structural constraints. This dual-stream framework enables high-quality novel-view reconstruction at low bitrates without scene-specific optimization or iterative diffusion sampling. The complete framework is optimized through three successive stages: feed-forward Gaussian pretraining, entropy-constrained Gaussian compression, and codec-aware generative adaptation.

\subsection{Detail-Aware Feed-Forward Gaussian Coding}
\label{sec:encoder}

%\subsection{Detail-Aware Feed-Forward Gaussian Coding}

%\subsection{Detail-Aware Feed-Forward Gaussian Coding}

%Here, we introduce a detail-aware feed-forward Gaussian coding framework that transforms sparse multi-view observations into a compact Gaussian structural stream through three complementary components. First, dual-branch detail-aware Gaussian prediction combines fine-grained local cues with local-global contextual aggregation to preserve informative appearance details while maintaining coherent geometry. Second, rate-distortion-guided Gaussian compaction removes redundant primitives according to their visual contribution relative to the estimated coding cost. Third, geometry-attribute decoupled coding separately compresses Gaussian centers and non-geometric attributes using dedicated geometry coding and hierarchical attribute entropy modeling. A lightweight reference appearance stream further provides complementary visual evidence that is difficult to retain in the Gaussian domain. Together, these components provide compact yet reliable structural and appearance cues for subsequent geometry-guided generative decoding. We detail each component below.

%\subsection{Detail-Aware Feed-Forward Gaussian Coding}

Here, we introduce a detail-aware feed-forward Gaussian coding framework that transforms sparse multi-view observations into a compact Gaussian structural stream through three complementary designs. First, a dual-branch detail-aware Gaussian prediction module combines local-global structural reasoning with fine-grained appearance features to produce coherent and detail-preserving Gaussian representations. Second, a rate-distortion-guided Gaussian compaction scheme removes redundant primitives according to their visual contribution relative to the estimated coding cost. Third, a geometry-attribute decoupled coding strategy separately compresses Gaussian centers and non-geometric attributes.  In parallel, a lightweight reference appearance stream preserves complementary appearance cues that are difficult to retain in compact Gaussian representations. Together, the two streams provide reliable structural anchors and compact appearance cues for subsequent geometry-guided generative decoding. %We detail each component below.

Given the input view set $\mathcal{I}$, the feed-forward Gaussian prediction network $\Phi_{\theta}$ directly infers a renderable Gaussian representation:
\begin{equation}
G
=
\Phi_{\theta}(\mathcal{I})
=
\{g_j\}_{j=1}^{M},
\qquad
g_j
=
\{
\boldsymbol{\mu}_j,
\mathbf{s}_j,
\mathbf{q}_j,
\alpha_j,
\mathbf{c}_j
\},
\label{eq:gaussian_prediction}
\end{equation}
where $M$ denotes the number of predicted Gaussian primitives. For the $j$-th Gaussian, $\boldsymbol{\mu}_j$, $\mathbf{s}_j$, $\mathbf{q}_j$, $\alpha_j$, and $\mathbf{c}_j$ denote its center, anisotropic scale, rotation quaternion, opacity, and RGB color, respectively.

Reliable Gaussian prediction from sparse observations requires both cross-view structural reasoning and fine-grained appearance modeling. However, deeply aggregated features used for correspondence reasoning may weaken local textures and high-frequency appearance cues. We therefore adopt two complementary feature pathways. The structural pathway aggregates local image evidence and cross-view context through a local-global attention backbone, providing coherent features for Gaussian prediction. In parallel, the detail pathway transforms intermediate features from a pretrained DINO backbone through a lightweight MLP to retain fine-grained appearance information.

The two pathways are fused asymmetrically. Gaussian centers are predicted from the structurally aggregated features, whereas the detail features are injected only into the Gaussian attribute prediction head. This design enhances fine-grained attribute prediction without disturbing cross-view center estimation. We further adopt zeroth-order spherical harmonics, such that each Gaussian carries a compact view-independent RGB vector rather than higher-order spherical harmonic coefficients.

To facilitate quantization, rate estimation, and learned attribute coding, we separate each Gaussian center from its remaining attributes and transform parameters with different constraints into coding-friendly domains:
\begin{equation}
\mathbf{a}_j
=
[
\mathbf{c}_j,
\operatorname{logit}(\alpha_j),
\log\mathbf{s}_j,
\mathbf{q}_j
]
\in\mathbb{R}^{11}.
\label{eq:attribute_parameterization}
\end{equation}
The logit transform maps bounded opacity values to an unconstrained domain, while the logarithmic transform reduces the dynamic range of positive scale parameters. During decoding, opacity and scale are recovered using the sigmoid and exponential functions, respectively, and the quaternion is normalized to satisfy the unit-norm constraint.

Although the predicted Gaussian set is directly renderable, feed-forward prediction typically generates dense primitives to ensure sufficient scene coverage. Directly encoding all primitives is inefficient because their visual contributions and coding costs vary substantially. We therefore introduce a rate-distortion-guided Gaussian compaction scheme, implemented by pruning primitives according to their visual contribution relative to the estimated coding cost.

For the $j$-th Gaussian, we estimate a visual contribution score $v_j$ from its opacity and spatial coverage and estimate its attribute coding cost ${b}_j^a$ using the learned entropy model. Since Gaussian centers are jointly compressed by the geometry coder, their coding cost cannot be reliably decomposed into primitive-level terms. We therefore use the estimated attribute cost as a practical proxy for the coding cost of each primitive. We perform RD-guided pruning using the following retention priority:
\begin{equation}
p_j^{\mathrm{RD}}
=
\frac{v_j}
{{b}_j^a+\epsilon},
\label{eq:rd_priority}
\end{equation}
where $\epsilon$ is a small constant for numerical stability. A larger $p_j^{\mathrm{RD}}$ indicates a higher visual return relative to the estimated attribute coding cost.

Given a target retention ratio, we rank all primitives according to $p_j^{\mathrm{RD}}$ and retain the top-ranked candidates, resulting in a compacted Gaussian set $G_S$, where $\mathcal{S}$ denotes the retained index set. This greedy pruning strategy removes primitives with low coding value while avoiding expensive combinatorial optimization.

After compaction, the retained Gaussian centers and attributes are compressed through separate coding paths. Gaussian centers form an irregular 3D point set with strong spatial occupancy correlations, whereas color, opacity, scale, and rotation exhibit heterogeneous distributions and spatial and channel dependencies. For geometry coding, the retained centers are first quantized and then compressed using G-PCC to produce the geometry bitstream $B_{\mu}$. This coding path exploits the spatial occupancy structure of the compacted Gaussian point set.

\begin{figure}[t]
    \centering
    \includegraphics[width=\linewidth]{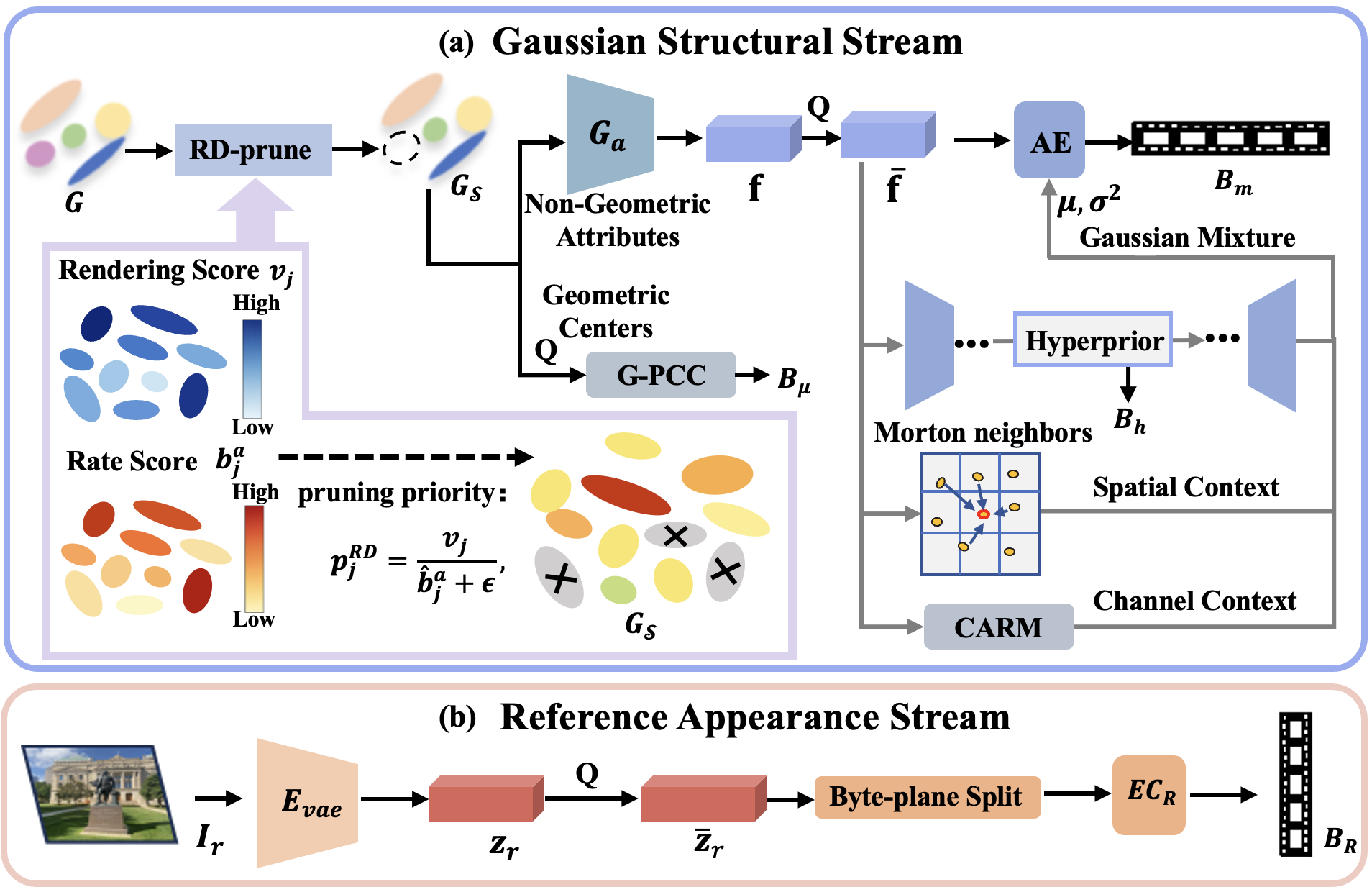}
    \caption{Architecture of the dual-stream coding scheme. 
(a) The Gaussian structural stream combines rate-distortion-guided pruning
with geometry-attribute decoupled coding to produce $B_G=\{B_{\mu},B_{\mathrm{m}},B_{\mathrm{h}}\}$. 
(b) The reference appearance stream quantizes and entropy-codes a selected reference image in the VAE latent domain to produce $B_R$. 
The two streams provide complementary structural and appearance information for generative decoding.}
    %\caption{\huber{Architecture of the proposed dual-stream feed-forward Gaussian encoder. (a) The Gaussian structural stream performs rate-distortion-aware primitive selection and separate geometry-attribute coding to produce $B_G$. (b) The reference appearance stream quantizes and entropy-codes a selected reference image in the VAE latent domain to produce $B_R$. Both streams are included in the total bitrate and provide complementary information for generative decoding.}}
    %Overview of the proposed R-D-aware feed-forward Gaussian encoder. The predicted Gaussian primitives are selected according to their visual contribution and estimated coding cost, after which the retained geometry and attributes are compressed through separate coding paths. A compressed reference appearance stream is additionally transmitted to provide appearance guidance for the generative decoder.}
    \label{fig:prune}
\end{figure}

For attribute coding, the retained Gaussians are first ordered according to the Morton codes of their quantized centers, such that spatially neighboring primitives tend to remain close in the resulting sequence. Their transformed attribute vectors are then arranged according to a predefined channel layout and mapped by the Gaussian analysis transform $G_a$ into compact latent features $\mathbf{f}$. The quantized latents $\bar{\mathbf{f}}$  are encoded using a hierarchical entropy model that combines hyperprior features, spatial context from previously decoded Morton neighbors, and channel context modeled by a channel-wise autoregressive model (CARM). The resulting attribute representation is encoded into the main attribute bitstream $B_{\mathrm{m}}$ and the hyperprior bitstream $B_{\mathrm{h}}$. The complete Gaussian structural stream is therefore
\begin{equation}
B_G
=
\{
B_{\mu},
B_{\mathrm{m}},
B_{\mathrm{h}}
\}.
\label{eq:gaussian_stream}
\end{equation}
By assigning dedicated coding mechanisms to Gaussian centers and attributes, the proposed geometry-attribute decoupled strategy better exploits their respective structural and statistical dependencies.

Although $B_G$ preserves reliable geometry, visibility, and coarse appearance, some high-frequency and view-dependent appearance information remains expensive to represent using compact Gaussian attributes. We therefore complement the Gaussian stream with a lightweight reference appearance stream. A set of candidate reference views is predefined, and the view closest to the query camera is selected according to camera proximity. Given the selected reference image $I_r$, its VAE latent is quantized and entropy-coded as
\begin{equation}
\bar{\boldsymbol{z}}_r
=
Q_r
\left(
E_{\mathrm{vae}}(I_r)
\right),
\qquad
B_R
=
EC_R
\left(
\bar{\boldsymbol{z}}_r
\right),
\label{eq:reference_stream}
\end{equation}
where $E_{\mathrm{vae}}(\cdot)$ denotes the VAE encoder, $Q_r(\cdot)$ denotes reference-latent quantization, $\bar{\boldsymbol{z}}_r$ is the quantized reference latent, and $EC_R(\cdot)$ denotes the corresponding entropy coder.

Before entropy coding, a byte-plane split operation reorganizes the quantized latent values according to byte significance to facilitate lossless coding. The bitrate of $B_R$ is explicitly included in the total rate and therefore introduces no unaccounted appearance side information. Consequently, $B_G$ provides a compact structural anchor, while $B_R$ supplies complementary appearance cues that are difficult to preserve in the Gaussian domain. The two codec-derived streams jointly condition the geometry-guided generative decoder described in the next subsection.

\subsection{Geometry-Guided One-Step Generative Decoding}
\label{subsec:generative_synthesis}

GenSplatCodec reconstructs the query view from the two transmitted streams through geometry-guided one-step generative decoding. The Gaussian stream provides a compact structural basis, while the reference stream supplies complementary appearance cues. Given the Gaussian stream $B_G$, the reference stream $B_R$, the query camera pose $\pi_q$, and the reference camera pose $\pi_r$, the decoder first recovers the compact Gaussian representation and the reference appearance latent:
%The decoder of GenSplatCodec adopts a geometry-guided one-step generative architecture to reconstruct the query view from the two transmitted streams produced by the encoder. Given the Gaussian stream $B_G$, the reference stream $B_R$, the query camera pose $\pi_q$, and the reference camera pose $\pi_r$, the decoder first recovers the compact Gaussian representation and the reference appearance latent: 
\begin{equation}                            
\hat{G}=Dec_G(B_G), \qquad                
\hat{\boldsymbol{z}}_r=ED_R(B_R),       
\end{equation}        

where $Dec_G(\cdot)$ denotes Gaussian-stream decoding, $\hat{G}$ is the decoded low-bitrate Gaussian representation, $ED_R(\cdot)$ denotes reference-stream entropy decoding, and $\hat{\boldsymbol{z}}_r$ is the decoded reference latent in the VAE latent domain. The decoded Gaussian representation provides a view-consistent 3D structural basis and coarse query-view appearance, while the decoded reference latent supplies complementary appearance information that is expensive to represent in the compact Gaussian stream.

Although $\hat{G}$ can be directly rendered at the query viewpoint, such rendering is insufficient under tight rate constraints because high-frequency textures, view-dependent appearance, and weakly observed details are severely attenuated by Gaussian pruning and quantization. We therefore use the decoded Gaussian representation as a structural anchor rather than the final reconstruction. Specifically, we render the query-view RGB anchor and the query/reference geometry cues from $\hat{G}$:       
\begin{equation}                                     (\hat{I}_q^g,\hat{D}_q^g)=\mathcal{R}(\hat{G},\pi_q), \qquad                                               \hat{D}_r^g=\mathcal{R}_d(\hat{G},\pi_r),            \end{equation}       
where $\mathcal{R}(\cdot)$ denotes Gaussian rendering that outputs both an RGB image and a depth-based geometry map, $\mathcal{R}_d(\cdot)$ denotes depth-only Gaussian rendering, $\hat{I}_q^g$ is the low-rate Gaussian rendering at the query view, $\hat{D}_q^g$ is the query-view geometry map, and $\hat{D}_r^g$ is the reference-view geometry map. The RGB anchor provides coarse photometric content for the desired query view, while the geometry maps describe the spatial layouts required for geometry-aware appearance transfer.          

The rendered query image and the decoded reference latent are then mapped to two coupled branches in the VAE latent space. The query branch is initialized by encoding the low-rate Gaussian rendering:            \begin{equation}                                     \boldsymbol{z}_q=E_{vae}(\hat{I}_q^g),            \end{equation}                                       where $\boldsymbol{z}_q$ is the query-view latent reconstructed from the Gaussian stream. The reference branch directly uses the decoded reference latent $\hat{\boldsymbol{z}}_r$ as its appearance condition. Thus, both branches are constructed entirely from the two transmitted bitstreams.

Directly transferring reference appearance across viewpoints is unreliable because differences in spatial layout, visibility, and occlusion may introduce texture misalignment and geometry-inconsistent details. Moreover, directly using rendered geometry maps as independent image-domain conditions is suboptimal because they are not naturally aligned with the latent representation of the pretrained generative model. We therefore realize geometry-guided one-step generative decoding through a hierarchical geometry control module that converts the rendered geometry cues into latent-aligned control signals.

%A key challenge is that reference appearance is generally insufficient for direct transfer to the query view, as differences in spatial layouts and visibility patterns between the two views may cause cross-view appearance interaction to introduce texture misalignment and geometry-inconsistent details in the query view. Directly using geometry maps as independent image-domain conditions is also suboptimal, since these cues are not naturally aligned with the latent distribution of the pretrained generative model. We therefore introduce geometry-guided one-step diffusion, implemented through a hierarchical geometry control module that converts decoder-rendered geometry cues into latent-aligned control signals for the one-step diffusion decoder. 

The view-specific geometry embeddings are first constructed as   
\begin{equation}                                     \mathbf{g}_q=\rho(\nu(\hat{D}_q^g)), \qquad          \mathbf{g}_r=\rho(\nu(\hat{D}_r^g)),                 \end{equation}                                       where $\nu(\cdot)$ normalizes the rendered depth geometry maps, $\rho(\cdot)$ maps them to the spatial resolution of the VAE latents, and $\mathbf{g}_q$ and $\mathbf{g}_r$ denote the query-view and reference-view geometry embeddings, respectively. A lightweight geometry controller then transforms each geometry embedding into multi-level control features:                                            \begin{equation}                                     \{\Delta\mathbf{H}_{v}^{\ell}\}_{\ell=1}^{L}=        C_{\phi}(\mathbf{g}_v), \qquad v\in\{q,r\},          \end{equation}                                       where $C_{\phi}(\cdot)$ is the geometry controller with parameters $\phi$, and $\Delta\mathbf{H}_{v}^{\ell}$ denotes the geometry control signal injected into the $\ell$-th latent stage of branch $v$. This design allows geometry cues at different spatial scales to constrain the latent processing hierarchy while remaining aligned with the latent generative process.

The query and reference appearance latents are processed as two coupled branches of a shared one-step multi-view generative U-Net. At each controlled latent stage, the corresponding geometry control signal is injected into the branch feature:          \begin{equation}                                     \mathbf{H}_{v}^{\ell}=U_{\omega}^{\ell}(\mathbf{H}_{v}^{\ell-1})+\Delta\mathbf{H}_{v}^{\ell}, \qquad v\in\{q,r\}, \end{equation}       
where $\mathbf{H}_{v}^{\ell}$ is the feature of branch $v$ at the $\ell$-th stage and $U_{\omega}^{\ell}(\cdot)$ denotes the $\ell$-th latent processing block of the one-step U-Net with parameters $\omega$. The initial branch features are initialized from the appearance latents, i.e., $\mathbf{H}_{q}^{0}=\boldsymbol{z}_q$ and $\mathbf{H}_{r}^{0}=\hat{\boldsymbol{z}}_r$. Within the attention layers, latent tokens from the query and reference branches are reshaped into a shared sequence and processed by the reference mixing layers, enabling cross-view latent interaction. The query branch can therefore aggregate complementary textures and appearance patterns from the reference branch, while the injected query- and reference-view geometry controls regulate how such information is interpreted and transferred, thereby reducing structural drift and spatially implausible detail synthesis.

The complete one-step generative decoding process is written as                                \begin{equation}                                     \tilde{\boldsymbol{z}}_q = \Pi_q\!\left[U_{\omega}^{1step} \left(               \boldsymbol{z}_q,\hat{\boldsymbol{z}}_r              \mid                                                 \{\Delta\mathbf{H}_{q}^{\ell}\}_{\ell=1}^{L},        \{\Delta\mathbf{H}_{r}^{\ell}\}_{\ell=1}^{L},        \mathbf{e}_p \right)\right],                         \end{equation}                                       where $U_{\omega}^{1step}(\cdot)$ denotes the geometry-controlled one-step multi-view generative U-Net, $\mathbf{e}_p$ is a scene-independent conditioning embedding inherited from the pretrained generative prior, $\Pi_q(\cdot)$ extracts the query-branch output, and $\tilde{\boldsymbol{z}}_q$ is the synthesized query latent. The final query image is reconstructed by the VAE decoder:                 
\begin{equation}                                     \tilde{I}_q=D_{vae}(\tilde{\boldsymbol{z}}_q),      \end{equation}                                       where $D_{vae}(\cdot)$ denotes the VAE decoder.      The proposed generative decoder serves as an integral decoding transform of the codec rather than an external image-domain post-processing module. The Gaussian stream determines the query-view structure, visibility, and coarse appearance, while the reference stream supplies compressed appearance cues. By coupling the two streams through hierarchical geometry control and cross-view latent interaction, the decoder reconstructs high-fidelity novel views in a single generative step, without iterative diffusion sampling, raw reference images, or unaccounted decoder-side information.

\subsection{ Three-Stage Optimization}

\label{subsec:training_inference}

The proposed framework is optimized through three successive stages that follow the dependency structure of the codec: feed-forward Gaussian representation learning, compact Gaussian coding, and geometry-guided generative decoding. Stage 1 learns a renderable detail-aware Gaussian representation without coding constraints. Stage 2 freezes the Gaussian predictor and optimizes rate-distortion-guided Gaussian compaction together with attribute quantization and hierarchical entropy modeling. Stage 3 fixes the Gaussian coding path and adapts the geometry-guided one-step generative decoder to the codec-derived structural and appearance conditions encountered during inference. This stage-wise optimization avoids unstable joint optimization of all components from scratch and ensures that each downstream module is optimized using the outputs produced by its actual upstream pathway.

\textbf{Stage 1: Feed-forward Gaussian pretraining.}  In the first stage, we train the detail-aware feed-forward Gaussian prediction network without Gaussian compaction, quantization, or rate constraints. Given sparse input views, the network predicts a Gaussian representation and renders a query view $\hat{I}_q$, where $q$ denotes the query viewpoint and $I_q$ is the corresponding ground-truth image. This stage is supervised by the rendering reconstruction objective:
\begin{equation}                                     \mathcal{L}_{ff}=\mathcal{L}_{mse}(\hat{I}_q,I_q)+\lambda_{lpips}\mathcal{L}_{LPIPS}(\hat{I}_q,I_q)+\lambda_{op}\mathcal{L}_{op},        \end{equation}                                       where $\mathcal{L}_{mse}$ is the pixel-wise mean squared error, $\mathcal{L}_{LPIPS}$ is the perceptual loss, and $\mathcal{L}_{op}$ is an opacity regularization term. Specifically,
\begin{equation}                                     \mathcal{L}_{op}=\frac{1}{M}\sum_{j=1}^{M}\alpha_j,  \end{equation}                                       where $M$ is the number of predicted Gaussian primitives and $\alpha_j$ is the opacity of the $j$-th Gaussian. The reconstruction terms encourage structurally coherent and perceptually faithful renderings, while $\mathcal{L}_{op}$ acts as an $\ell_1$-style sparsity prior on primitive opacity, encouraging unnecessary semi-transparent primitives to become inactive.  This provides a more compression-friendly initialization for the subsequent Gaussian coding stage.

\textbf{Stage 2: Entropy-constrained Gaussian compression.} In the second stage, we introduce the rate-distortion-guided Gaussian compaction, attribute quantization, and hierarchical attribute entropy modeling described in Sec.~\ref{sec:encoder}. The pretrained feed-forward Gaussian predictor remains fixed and provides the upstream Gaussian representation for compression optimization. This stage aims to reduce primitive redundancy and attribute coding cost while preserving the geometry and coarse appearance required by the subsequent generative decoder. Let $\hat{I}_q^g$ denote the query view rendered from the compressed and reconstructed Gaussian representation. The compression path is optimized by

\begin{equation}         \mathcal{L}_{codec}=\mathcal{L}_{fid}(\hat{I}_q^g,I_q)+\lambda_{rate}\mathcal{L}_{rate},  
\end{equation}                           where $\mathcal{L}_{fid}$ measures the rendering fidelity of the reconstructed Gaussian representation and $\mathcal{L}_{rate}$ estimates the normalized coding cost of the learned attribute latents. The fidelity term is defined as     
\begin{equation}                                     \mathcal{L}_{fid}=\lambda_{mse}\|\hat{I}_q^g-I_q\|_2^2+\lambda_{ssim}\left[1-SSIM(\hat{I}_q^g,I_q)\right],                        \end{equation}                                       where the MSE term constrains color and content consistency, while  the SSIM term preserves local luminance, contrast, and structural similarity. The differentiable rate term is computed from the estimated entropy of the main attribute latent and the hyperprior latent: \begin{equation}                                     \mathcal{L}_{rate}=\frac{R_m+R_h}{M_rD_a},           \end{equation}                                      
where $R_m$ and $R_h$ denote the estimated rates of the main and hyperprior latents, respectively, $M_r$ is the number of retained Gaussians after pruning, and $D_a$ is the number of coded attributes per Gaussian. This normalized rate term encourages compact latent distributions and efficient attribute entropy coding, while the rate-distortion-guided compaction controls the number and composition of retained primitives according to their visual contribution relative to the estimated attribute cost. 

%After this stage, the structural coding path is fixed and used to generate codec-consistent inputs for generative adaptation.  

Since the Gaussian centers are encoded using the non-differentiable G-PCC codec, the geometry bitstream $B_{\mu}$ is not included in $\mathcal{L}_{rate}$. Its coding cost is instead reduced indirectly through Gaussian compaction. The reference appearance stream is also not part of this Stage-2 attribute-rate surrogate. Nevertheless, the actual sizes of the geometry, attribute, hyperprior, and reference bitstreams are all included in the total rate reported during evaluation. After this stage, the Gaussian coding path is fixed and used to generate codec-derived structural conditions for generative adaptation.

\textbf{Stage 3: Codec-aware generative adaptation.}  In the third stage, we train the geometry-guided one-step generative decoder with the complete coding path for dual-stream codec-aware adaptation. The feed-forward Gaussian predictor and compression modules remain fixed, while the generative components, including the hierarchical geometry controller and the one-step multi-view U-Net, are optimized within the complete decoding pathway. For each training sample, the decoded Gaussian representation is rendered at the query and reference viewpoints to produce the low-rate RGB anchor and geometry cues, and the reference stream provides the coded appearance latent. These signals follow the same reconstruction pathway used during inference, allowing the generative decoder to adapt to the structural and appearance information actually retained by the two coded streams rather than to idealized uncompressed inputs. The reconstructed query view $\tilde{I}_q$ is supervised by a fidelity-perception objective:
\begin{equation}                                     \mathcal{L}_{gen}=\lambda_2\|\tilde{I}_q-{I}_q\|_2^2+\lambda_{lpips}\mathcal{L}_{LPIPS}(\tilde{I}_q,I_q),                                   \end{equation}   
where the pixel-domain term preserves query-view content and low-frequency structure, while the perceptual term promotes visually faithful high-frequency details that are difficult to transmit explicitly.

At inference, GenSplatCodec performs feed-forward encoding and decoding without scene-specific optimization. The encoder produces the Gaussian stream $B_G$ and the reference stream $B_R$, and the decoder reconstructs novel views through Gaussian decoding, geometry rendering, reference-latent decoding, hierarchical geometry control, and one-step generative decoding. The reported bitrate includes both transmitted streams:          
\begin{equation}                                     R_{total}=R_G+R_R,                                   \end{equation}                                       where $R_G$ is the bitrate of the Gaussian stream and $R_R$ is the bitrate of the reference stream. Thus, the decoder does not rely on raw reference images or unaccounted side information, and the final reconstruction is obtained in a single feed-forward pass without scene-specific optimization or iterative diffusion sampling.

\section{Experiments}
\label{sec:experiments}

\subsection{Experimental Settings}
\label{subsec:exp_settings}

\subsubsection{Datasets}
We evaluate GenSplatCodec on two standard benchmarks for feed-forward 3D Gaussian splatting: DL3DV~\cite{ling2024dl3dv} and RealEstate10K (RE10K)~\cite{zhou2018stereo}. DL3DV contains diverse indoor and outdoor scenes with challenging reflections, transparency, and illumination variations. Following commonly adopted feed-forward evaluation protocols, we evaluate at resolutions of $224\times224$ and $518\times518$, using $2$, $4$, and $6$ context views to assess robustness to different levels of input sparsity. RE10K mainly consists of indoor and residential scenes collected from online videos, for which we adopt the standard two-context-view protocol at $224\times224$ resolution. All methods are evaluated on the official or widely adopted test splits without scene-specific optimization.

%We evaluate GenSplatCodec on two standard benchmarks for feed-forward 3D Gaussian splatting, namely DL3DV\cite{ling2024dl3dv} and RealEstate10K (RE10K)\cite{zhou2018stereo}. RealEstate10K mainly contains multi-view indoor and residential scenes collected from online videos, while DL3DV is a large-scale novel-view synthesis dataset covering diverse indoor and outdoor environments with challenging reflections, transparency, and illumination variations. For DL3DV, we follow the commonly adopted feed-forward evaluation protocols at resolutions of $224\times224$ and $518\times518$. We consider $2$, $4$, and $6$ context views to evaluate the performance under different levels of input sparsity. For RE10K, we adopt the standard two-context-view protocol at a resolution of $224\times224$. All methods are evaluated on the official or commonly used benchmark test splits without scene-specific optimization.

\begin{table*}[!ht]
\centering
\caption{Quantitative comparison on the DL3DV and RealEstate10K datasets at $224\times224$ resolution. Size denotes the average total bitstream size per scene. The best and second-best results are highlighted in bold and underlined, respectively.}
\label{tab:lowres_main}
%\footnotesize
\fontsize{10pt}{10pt}\selectfont
\renewcommand{\arraystretch}{0.85}
\setlength{\tabcolsep}{5.0pt}
% \resizebox{\textwidth}{!}{
% \begin{tabular}{c|l|ccc|ccc|c}
\begin{tabular*}{\textwidth}{
@{\extracolsep{\fill}}
c|l|ccc|ccc|c
@{}
}
\toprule
\multirow{2}{*}{Dataset} &
\multirow{2}{*}{Method} &
\multicolumn{3}{c|}{Full-Reference Fidelity} &
\multicolumn{3}{c|}{Perceptual Quality} &
\multirow{2}{*}{\makecell{Size\\(MB)$\downarrow$}} \\
\cmidrule(lr){3-5}\cmidrule(lr){6-8}
& & PSNR$\uparrow$ & SSIM$\uparrow$ & LPIPS$\downarrow$ & FID$\downarrow$ & MUSIQ$\uparrow$ & MANIQA$\uparrow$ & \\
\midrule

\multirow{12}{*}{DL3DV}
& DepthSplat      
& 19.35 & 0.599 & 0.272 & 57.85 & \underline{46.60} & \underline{0.245} & 20.500 \\

& DepthSplat+SOGS 
& 19.36 & 0.598 & 0.276 & 59.24 & 46.26 & 0.242 & 5.622 \\

& DepthSplat+FCGS 
& 18.23 & 0.538 & 0.415 & 152.41 & 33.55 & 0.161 & 1.760 \\

& AnySplat    
& 12.67 & 0.219 & 0.413 & 97.56 & 43.04 & 0.230 & 32.980 \\

& AnySplat+SOGS   
& 12.70 & 0.211 & 0.414 & 100.36 & 43.14 & 0.224 & 3.578 \\

& AnySplat+FCGS   
& 12.63 & 0.237 & 0.419 & 107.68 & 43.08 & 0.227 & 1.118 \\

& YoNoSplat       
& \underline{20.09} & 0.617 & \underline{0.201} & 54.51 & 43.16 & 0.224 & 6.508 \\

& YoNoSplat+SOGS  
& 20.04 & 0.614 & 0.213 & 58.03 & 41.20 & 0.207 & 1.199 \\

& YoNoSplat+FCGS  
& 18.55 & 0.583 & 0.360 & 87.89 & 38.62 & 0.202 & 1.457 \\

& Ours-Low        
& 19.21 & 0.562 & 0.260 & 65.30 & 44.49 & 0.214 & \textbf{0.206} \\

& Ours-Mid        
& 20.02 & \underline{0.619} & 0.209 & \underline{53.20} & 43.69 & 0.222 & \underline{0.376} \\

& Ours-High       
& \textbf{20.76} & \textbf{0.638} & \textbf{0.184} & \textbf{50.51} & \textbf{47.96} & \textbf{0.249} & 0.830 \\

\midrule

\multirow{12}{*}{RealEstate10K}

& DepthSplat      
& 22.54 & \underline{0.800} & 0.181 & \underline{35.89} & 45.44 & \underline{0.256} & 15.690 \\

& DepthSplat+SOGS 
& 22.48 & 0.796 & 0.187 & 37.45 & 45.23 & 0.252 & 3.609 \\

& DepthSplat+FCGS 
& 22.07 & \textbf{0.805} & 0.221 & 49.27 & 40.79 & 0.224 & 1.700 \\

& AnySplat        
& 14.55 & 0.415 & 0.335 & 70.33 & 42.22 & 0.230 & 30.850 \\

& AnySplat+SOGS   
& 14.57 & 0.414 & 0.337 & 75.74 & 41.97 & 0.221 & 2.913 \\

& AnySplat+FCGS   
& 14.50 & 0.448 & 0.351 & 89.04 & 42.38 & 0.217 & 1.045 \\

& YoNoSplat       
& \underline{23.31} & 0.785 & \underline{0.136} & 36.39 & 42.27 & 0.240 & 6.508 \\

& YoNoSplat+SOGS  
& 23.25 & 0.783 & 0.141 & 37.16 & 41.36 & 0.229 & 1.246 \\

& YoNoSplat+FCGS  
& 21.63 & 0.767 & 0.257 & 50.17 & 40.89 & 0.224 & 1.400 \\

& Ours-Low        
& 21.27 & 0.732 & 0.190 & 47.93 & \underline{45.81} & 0.247 & \textbf{0.244} \\

& Ours-Mid        
& 22.46 & 0.761 & 0.150 & 39.20 & 45.20 & \underline{0.256} & \underline{0.602} \\

& Ours-High       
& \textbf{23.33} & 0.791 & \textbf{0.125} & \textbf{32.45} & \textbf{45.83} & \textbf{0.272} & 0.748 \\

\bottomrule
\end{tabular*}
\end{table*}

\subsubsection{Baselines}
We compare GenSplatCodec with two groups of baselines. The first group includes representative feed-forward Gaussian reconstruction methods, namely DepthSplat~\cite{xu2025depthsplat}, AnySplat~\cite{jiang2025anysplat}, and YoNoSplat~\cite{ye2025yonosplat}, whose original Gaussian representations serve as uncompressed references for reconstruction quality and storage cost. The second group consists of cascaded compression baselines that combine these reconstruction methods with existing 3DGS compressors. Specifically, we export the full Gaussian representation predicted by each reconstruction model and compress it using FCGS~\cite{chen2025fast} or SOGS~\cite{morgenstern2024compact}. For FCGS, we use the released model with $\lambda_{\mathrm{FCGS}}=0.0001$; for SOGS, we adopt its default compression configuration. In the tables, ``A+B'' denotes a cascaded pipeline in which method A predicts the Gaussian representation and method B subsequently compresses it.

% We compare GenSplatCodec with two groups of baselines. The first group consists of representative open-source feed-forward Gaussian reconstruction methods, including
% DepthSplat\cite{xu2025depthsplat}, AnySplat\cite{jiang2025anysplat}, and YoNoSplat\cite{ye2025yonosplat}. These methods serve as uncompressed reconstruction references and characterize the rendering quality and storage cost of the original feed-forward Gaussian representations. The second group consists of optimization-free compression pipelines constructed by combining the above feed-forward reconstruction methods with existing 3DGS compressors. Specifically, we first export the complete Gaussian primitives predicted by each feed-forward model and then directly apply FCGS\cite{chen2025fast} or SOGS\cite{morgenstern2024compact} to the resulting Gaussian representation. FCGS performs learned compression of the generated Gaussian attributes, for which we use the
% released model with $\lambda_{\mathrm{FCGS}}=0.0001$. SOGS
% reorganizes Gaussian parameters into locally smooth 2D grids before coding, and we adopt its default compression configuration. In the tables, ``A+B'' denotes a cascaded baseline in which method A first predicts the Gaussian primitives and method B subsequently compresses the exported representation.

\subsubsection{Evaluation Metrics}
We use established metrics to evaluate the rate-distortion-perception performance of GenSplatCodec. Reconstruction fidelity is measured using PSNR and SSIM~\cite{wang2004image}, with higher values indicating better agreement with the ground truth. LPIPS~\cite{zhang2018unreasonable} measures perceptual similarity, with lower values indicating better performance. We further report FID~\cite{heusel2017gans} to assess distributional realism, and MUSIQ~\cite{ke2021musiq} and MANIQA~\cite{yang2022maniqa} to evaluate no-reference perceptual quality. Compression efficiency is measured by the average total bitstream size per scene in MB, including both the Gaussian structural stream and the reference appearance stream.

%We adopted established metrics to evaluate the rate-distortion-perception performance of GenSplatCodec. Reconstruction fidelity is measured using PSNR and SSIM\cite{wang2004image}, where higher values indicate better pixel-level and structural agreement with the ground truth. Perceptual similarity is evaluated using LPIPS\cite{zhang2018unreasonable}, with a lower value indicating a better perceptual correspondence. We also report FID\cite{heusel2017gans}, MUSIQ\cite{ke2021musiq} and MANIQA\cite{yang2022maniqa} to evaluate visual realism and no-reference perceptual quality of the reconstructed views. Compression efficiency is measured by the average bitstream size per scene in MB.

\subsection{Comparison}
\label{subsec:main_comparison}

\textbf{Quantitative Comparisons.} Table~\ref{tab:lowres_main} compares GenSplatCodec with feed-forward Gaussian reconstruction methods and cascaded compression
methods on DL3DV and RealEstate10K at $224\times224$ resolution. GenSplatCodec consistently achieves superior quality-storage
trade-offs on both datasets.

On DL3DV, Ours-Low achieves 19.21 dB PSNR with only 0.206 MB
per scene, requiring approximately \textbf{1\%} of the storage used by
DepthSplat, which obtains a comparable PSNR of 19.35 dB with
20.500 MB. Compared with YoNoSplat, Ours-Mid maintains comparable
reconstruction fidelity, with 20.02 versus 20.09 dB PSNR, while
reducing the representation size from 6.508 to 0.376 MB, corresponding
to a $\boldsymbol{17.3\times}$ reduction in storage.
At a similar PSNR to YoNoSplat+SOGS (20.02 versus 20.04 dB),
Ours-Mid uses 68.6\% less storage while achieving better SSIM,
LPIPS, and FID. At the high-rate operating point, Ours-High achieves
the best results across all six reported quality metrics on DL3DV,
including 20.76 dB PSNR, 0.638 SSIM, 0.184 LPIPS, and 50.51 FID,
with only 0.830 MB per scene.

On RealEstate10K, Ours-High achieves the highest PSNR of 23.33 dB,
the lowest LPIPS of 0.125 and FID of 32.45, and the highest MUSIQ
and MANIQA scores of 45.83 and 0.272, respectively. Compared with
YoNoSplat, it improves all reported quality metrics while reducing
the representation size from 6.508 to 0.748 MB, corresponding to an
\textbf{88.5\%} reduction. These
results demonstrate that GenSplatCodec preserves high reconstruction
fidelity and perceptual quality under tight storage budgets through
compact Gaussian coding and geometry-guided generative decoding.

\begin{figure*}[t]
    \centering
    \includegraphics[width=\linewidth]{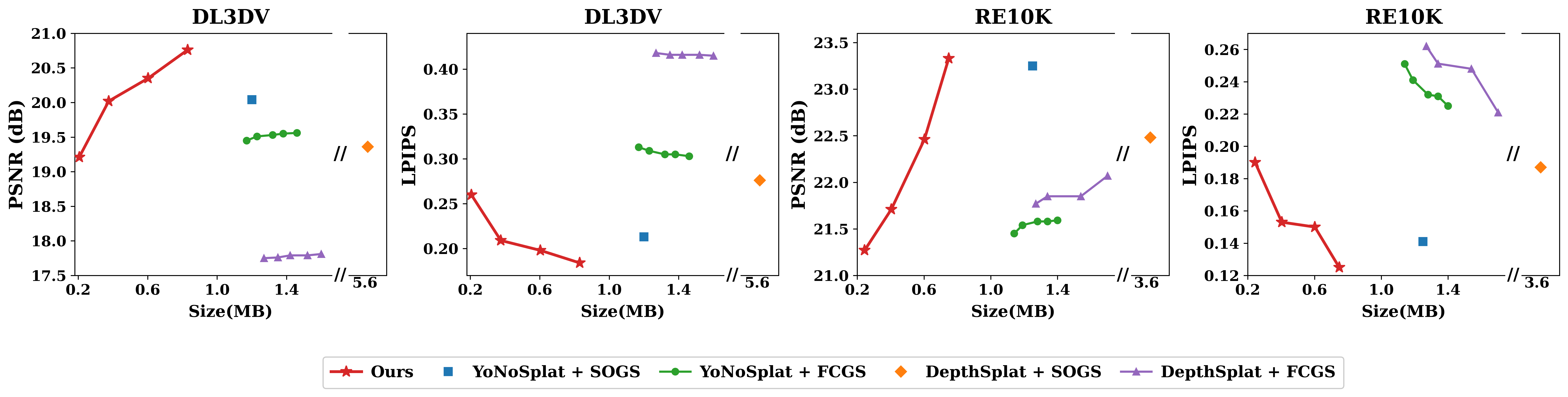}
    \caption{
    Rate-distortion curves on the DL3DV and RealEstate10K datasets.
The RD curves demonstrate the superiority of GenSplatCodec over existing Gaussian compression pipelines and highlight its effectiveness in achieving high reconstruction quality under extremely limited storage budgets.
    }
    \label{fig:rd_curve}
\end{figure*}

\textbf{Rate-Distortion Curves.}
Fig.~\ref{fig:rd_curve} compares the rate-distortion performance of
GenSplatCodec with representative cascaded Gaussian compression
methods on DL3DV and RealEstate10K. Across the evaluated operating
points, GenSplatCodec consistently achieves more favorable
PSNR-size and LPIPS-size trade-offs, with particularly pronounced
advantages under tight storage budgets. These results indicate that
jointly designing compact structural coding and geometry-guided
generative reconstruction better preserves reconstruction fidelity and
perceptual quality than cascading existing 3DGS compressors after
feed-forward Gaussian reconstruction.

% Fig.~\ref{fig:rd_curve} presents the RD curves of GenSplatCodec compared with representative Gaussian compression pipelines on the DL3DV and RealEstate10K datasets. 
% The RD curves clearly show that GenSplatCodec achieves superior RD performance across a wide range of bitrate levels, particularly in the low-bitrate regime where existing compression methods suffer from significant quality degradation. 
% Compared with YoNoSplat+SOGS, GenSplatCodec achieves comparable reconstruction quality while reducing the storage cost by \textbf{68.6}\% and \textbf{40.0}\% on the DL3DV and RealEstate10K datasets, respectively. 
% Moreover, GenSplatCodec exhibits a steeper RD curve slope, indicating more efficient utilization of additional bitrate for improving reconstruction fidelity. 
% These results demonstrate the superior rate-distortion efficiency of GenSplatCodec by jointly optimizing compact Gaussian transmission and geometry-guided generative reconstruction.

\begin{table}[t]
\centering
\caption{Compression performance under different numbers of input views on DL3DV. Size denotes the average storage cost per scene in MB.}
\label{tab:num_views}
\scriptsize
\setlength{\tabcolsep}{3.2pt}
\renewcommand{\arraystretch}{1.08}
\resizebox{\columnwidth}{!}{
\begin{tabular}{c|l|ccccc}
\toprule
Views & Method & PSNR$\uparrow$ & SSIM$\uparrow$ & LPIPS$\downarrow$ & Size$\downarrow$ & FID$\downarrow$ \\
\midrule

\multirow{5}{*}{2}
& DepthSplat+SOGS & 19.36 & 0.598 & 0.276 & 5.622 & 59.24 \\
& DepthSplat+FCGS & 18.23 & 0.538 & 0.415 & 1.760 & 152.41 \\
& YoNoSplat+SOGS  & \textbf{20.04} & \underline{0.614} & \underline{0.213} & \underline{1.199} & \underline{58.03} \\
& YoNoSplat+FCGS  & 18.55 & 0.583 & 0.360 & 1.457 & 87.89 \\
& Ours            & \underline{20.02} & \textbf{0.619} & \textbf{0.209} & \textbf{0.376} & \textbf{53.20} \\
\midrule

\multirow{5}{*}{4}
& DepthSplat+SOGS & 23.74 & \underline{0.803} & 0.137 & 11.420 & \underline{23.48} \\
& DepthSplat+FCGS & 20.80 & 0.699 & 0.292 & 2.610 & 88.05 \\
& YoNoSplat+SOGS  & \underline{24.24} & 0.802 & \underline{0.111} & \underline{1.941} & 24.86 \\
& YoNoSplat+FCGS  & 21.85 & 0.761 & 0.233 & 2.321 & 46.12 \\
& Ours            & \textbf{24.26} & \textbf{ 0.808} & \textbf{0.103} & \textbf{0.989} & \textbf{22.89} \\
\midrule

\multirow{5}{*}{6}
& DepthSplat+SOGS & 23.04 & 0.780 & 0.153 & 17.160 & \underline{23.66} \\
& DepthSplat+FCGS & 20.45 & 0.682 & 0.299 & 4.287 & 82.27 \\
& YoNoSplat+SOGS  & \underline{23.76} & \underline{0.792} & \underline{0.113} & \underline{2.830} & 23.81 \\
& YoNoSplat+FCGS  & 22.58 & 0.756 & 0.197 & 6.282 & 34.40 \\
& Ours            & \textbf{23.83} & \textbf{0.794} & \textbf{0.108} & \textbf{1.089} & \textbf{22.56} \\
\bottomrule

\end{tabular}}
\end{table}

\textbf{Robustness to Input Sparsity.}
Table~\ref{tab:num_views} evaluates GenSplatCodec with $2$, $4$, and
$6$ context views on DL3DV. Across all input settings, GenSplatCodec
consistently achieves the smallest average bitstream size while
maintaining the best or near-best reconstruction fidelity and
perceptual quality. Under the two-view setting, our method requires
only 0.376 MB per scene and achieves the best SSIM, LPIPS, and FID,
with only a 0.02 dB PSNR difference from YoNoSplat+SOGS. With four
and six context views, GenSplatCodec achieves the best results across
all reported quality metrics using only 0.989 and 1.089 MB per scene,
respectively. These results demonstrate robust quality-storage
performance across different levels of input sparsity.

\begin{table}[h]
\centering
\caption{Quantitative evaluation of multi-view reconstruction consistency on the DL3DV dataset.}
\label{tab:mv_consistency}
\scriptsize
\setlength{\tabcolsep}{4.0pt}
\renewcommand{\arraystretch}{1.08}
\resizebox{\columnwidth}{!}{
\begin{tabular}{l|ccc|ccc}
\toprule
Method &
SIFT$\uparrow$ & CLIP$\uparrow$ & LPIPS$\downarrow$ &
MS$\uparrow$ & BC$\uparrow$ & SC$\uparrow$ \\
\midrule
MVSplat360 
& 78.90 & 0.938 & 0.338 
& 0.891 & 0.932 & 0.868 \\

LatentSplat 
& 74.11 & 0.926 & 0.365 
& 0.873 & 0.919 & 0.867 \\

Ours       
& \textbf{90.88} & \textbf{0.941} & \textbf{0.332} 
& \textbf{0.897} & \textbf{0.935} & \textbf{0.878} \\

\bottomrule
\end{tabular}}
\end{table}

\textbf{Multi-View Consistency.}
Table~\ref{tab:mv_consistency} evaluates the consistency of generated novel views on DL3DV. We compare GenSplatCodec with MVSplat360~\cite{chen2024mvsplat360} and LatentSplat~\cite{wewer2024latentsplat}, both of which refine or generate novel-view renderings. SIFT correspondences, CLIP similarity, and cross-view LPIPS measure pairwise correspondence, while the VBench metrics (e.g., MS, BC, SC)~\cite{huang2024vbench,zheng2025vbench,huang2025vbench++} evaluate scene-level consistency across generated views.

GenSplatCodec achieves the best performance across all evaluated metrics. In particular, it obtains 90.88 SIFT correspondences, improving over the strongest baseline by 15.2\%, while achieving the lowest cross-view LPIPS of 0.332. It also obtains the highest scores on all three VBench consistency metrics. These results indicate that hierarchical geometry control effectively constrains appearance transfer and reduces view-inconsistent generation.

\textbf{High-Resolution Evaluation.}
Table~\ref{tab:highres_main}  reports the results on DL3DV at $518\times518$ resolution. GenSplatCodec achieves the best overall performance among the evaluated compression pipelines, with 18.71 dB PSNR, 0.332 LPIPS, and 67.95 FID at only 0.577 MB per scene. Compared with DepthSplat+SOGS, GenSplatCodec slightly improves PSNR from 18.65 to 18.71 dB and reduces LPIPS from 0.372 to 0.332, while decreasing the representation size from 19.741 to 0.577 MB, corresponding to a $34.2\times$ reduction. The results confirm that the proposed codec remains effective when scaling to higher rendering resolutions.

% \begin{table}[h]
% \centering
% \caption{Analysis of bitstream allocation under different quantization parameters on the DL3DV dataset.}
% \label{tab:quant}
% \scriptsize
% \setlength{\tabcolsep}{6.0pt}
% \renewcommand{\arraystretch}{1.08}
% \resizebox{0.95\columnwidth}{!}{
% \begin{tabular}{c|cccc|c}
% \toprule
% \multirow{2}{*}{\lambda_{rate}} &
% \multicolumn{4}{c|}{Bitstream Proportion (\%)} &
% \multirow{2}{*}{\makecell{Total Size\\(MB)$\downarrow$}} \\
% \cmidrule(lr){2-5}
% & $B_u$ & $B_m$ & $B_h$ & $B_R$ & \\
% \midrule
% $0.002$ & 0.2239 & 0.603 & 0.0027 & 0.0054 & 0.830 \\
% $0.050$  & 0.1117  & 0.3758 & 0.0027 & 0.0054 & 0.602 \\
% $0.100$   & 0.0720  & 0.2583 & 0.0021 & 0.0054 & 0.376 \\
% $1.000$   & 0.0720  & 0.1658 & 0.0021 & 0.0054 & 0.206 \\
% \bottomrule
% \end{tabular}}
% \end{table}

\begin{figure*}[t]
    \centering
    \includegraphics[width=\linewidth]{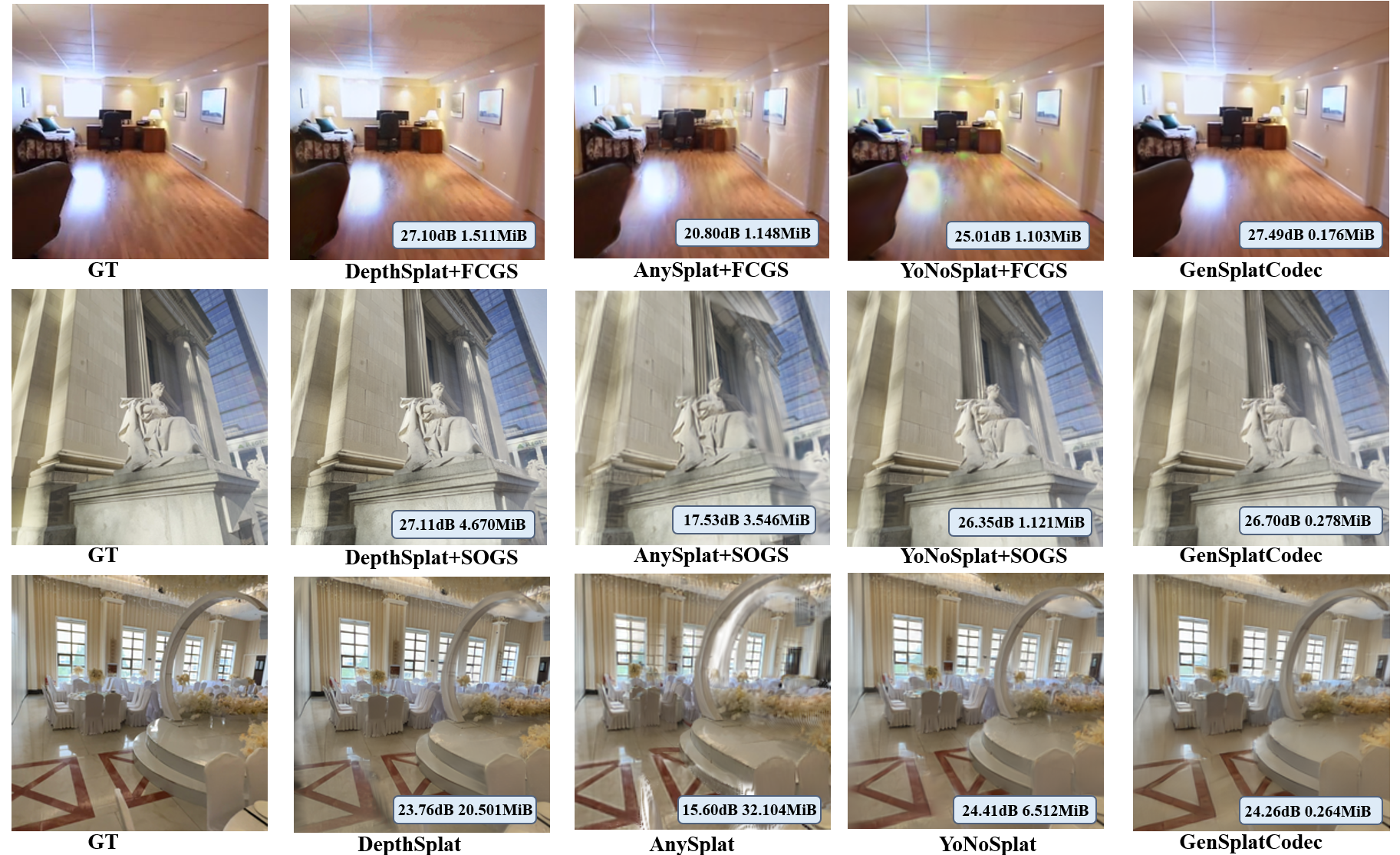}
    \caption{
    Qualitative comparison of novel-view reconstruction results. 
    Existing Gaussian compression methods suffer from noticeable texture degradation and rendering artifacts under aggressive compression. 
    In contrast, GenSplatCodec preserves structural consistency and recovers high-frequency details with significantly lower storage cost through geometry-guided one-step generative decoding.
    }
    \label{fig:visual}
\end{figure*}

\textbf{Qualitative Comparison.}
Fig.~\ref{fig:visual} compares GenSplatCodec with representative feed-forward Gaussian reconstruction methods and their compressed variants. Despite using substantially smaller representations, GenSplatCodec preserves sharper textures and more coherent scene structures under aggressive compression. In the indoor examples, our method retains floor and wall textures while reducing the over-smoothing artifacts observed in the compressed baselines. In scenes containing complex geometry, such as statues and curved architectural structures, GenSplatCodec better preserves object contours and boundary details. These comparisons show that geometry-guided generative decoding recovers perceptually important details without introducing noticeable structural inconsistencies.

\begin{table}[t]
\centering
\caption{Novel-view reconstruction and compression performance on DL3DV at $518\times518$ resolution.}
\label{tab:highres_main}
\scriptsize
\setlength{\tabcolsep}{5.0pt}
\renewcommand{\arraystretch}{1.1}
\resizebox{\columnwidth}{!}{
\begin{tabular}{l|ccc|c}
\toprule
Method &
PSNR$\uparrow$ &
LPIPS$\downarrow$ &
FID$\downarrow$ &
\makecell{Size(MB)$\downarrow$} \\
\midrule

DepthSplat+SOGS 
& \underline{18.65} & 0.372 & \underline{68.02} & 19.741 \\

DepthSplat+FCGS 
& 16.75 & \underline{0.343} & 176.97 & 6.732 \\

AnySplat+SOGS   
& 13.93 & 0.466 & 74.35 & 8.798 \\

AnySplat+FCGS   
& 13.84 & 0.371 & 78.23 & 4.728 \\

YoNoSplat+SOGS  
& 17.83 & 0.431 & 94.53 & 1.905 \\

YoNoSplat+FCGS  
& 17.17 & 0.493 & 111.96 & \underline{1.000} \\

Ours       
& \textbf{18.71} & \textbf{0.332} & \textbf{67.95} & \textbf{0.577} \\

\bottomrule
\end{tabular}}
\end{table}

% \textbf{Qualitative Comparison.}
% We present qualitative comparisons with representative feed-forward Gaussian reconstruction and compression pipelines, including DepthSplat\cite{xu2025depthsplat}, AnySplat\cite{jiang2025anysplat}, YoNoSplat\cite{ye2025yonosplat}, and their compressed variants, as shown in Figure~\ref{fig:visual}. 
% GenSplatCodec achieves comparable reconstruction quality to high-quality feed-forward Gaussian representations while operating at substantially lower storage costs. 
% Compared with existing compression pipelines, our method more effectively preserves fine appearance details and coherent scene structures under aggressive compression. 
% Specifically, GenSplatCodec maintains sharp textures on planar regions, such as floor patterns and wall appearances, while avoiding the over-smoothing artifacts observed in compressed baselines. 
% For complex geometric regions, including statues and curved structures, our method preserves clearer object contours and boundary details, whereas competing approaches suffer from structural degradation and detail loss. 
% Moreover, GenSplatCodec reconstructs visually consistent appearances in challenging indoor scenes with only a compact representation size, achieving perceptual quality comparable to uncompressed feed-forward models. 
% These results demonstrate that GenSplatCodec effectively combines compact Gaussian transmission with geometry-guided generative reconstruction, enabling realistic novel-view synthesis under extremely limited bitrate constraints.

\begin{table*}[t]
\centering
\caption{Ablation study validating the effectiveness of individual components on the DL3DV dataset.}
\label{tab:ablation}
\footnotesize
\setlength{\tabcolsep}{5.0pt}
\renewcommand{\arraystretch}{1.12}
\resizebox{0.9\textwidth}{!}{
\begin{tabular}{l|ccc|ccc|c}
\toprule
\multirow{2}{*}{Variant} &
\multicolumn{3}{c|}{Full-Reference Fidelity} &
\multicolumn{3}{c|}{Perceptual Quality} &
\multirow{2}{*}{\makecell{Size\\(MB)$\downarrow$}} \\
\cmidrule(lr){2-4}\cmidrule(lr){5-7}
& PSNR$\uparrow$ & SSIM$\uparrow$ & LPIPS$\downarrow$ & FID$\downarrow$ & MUSIQ$\uparrow$ & MANIQA$\uparrow$ & \\
\midrule
Base  
& 18.09 & 0.508 & 0.411 & 77.94 & 61.50 & 0.216 & 15.610 \\

+ Dual-Branch Gaussian Prediction
& 18.21 & 0.515 & 0.407 & 85.35 & 60.42 & 0.213 & 15.784 \\

+ Geometry–Attribute Decoupled Coding
& 17.91 & 0.521 & 0.470 & 111.96 & 40.07 & 0.135 & 4.679 \\

+ RD-Guided Gaussian Compaction
& 17.87 & 0.524 & 0.478 & 111.84 & 41.77 & 0.134 & \textbf{0.577} \\

+  Naive One-Step SD 
& 17.76 & 0.501 & 0.343 & 79.32 & 56.42 & 0.264 & \textbf{0.577} \\

+ Dual-Stream Codec-Aware Adaptation 
& 18.57 & 0.545 & 0.334 & 74.31 & 64.17 & 0.279 & \textbf{0.577} \\

+ Geometry-Guided SD 
& \textbf{18.71} & \textbf{0.553} & \textbf{0.332} & \textbf{67.95} & \textbf{65.66} & \textbf{0.281} & \textbf{0.577} \\

\bottomrule
\end{tabular}}
\end{table*}

\subsection{ Evaluation}

\begin{table}[h]
\centering
\caption{Analysis of bit allocation under different rate parameters on the DL3DV dataset.}
\label{tab:quant}
\footnotesize
\setlength{\tabcolsep}{5.0pt}
\renewcommand{\arraystretch}{1.12}
\resizebox{0.95\columnwidth}{!}{
\begin{tabular}{c|cccc|c}
\toprule
\multirow{2}{*}{$\lambda_{\mathrm{rate}}$} &
\multicolumn{4}{c|}{Bitstream Proportion} &
\multirow{2}{*}{\makecell{Size\\(MB)$\downarrow$}} \\
\cmidrule(lr){2-5}
& $B_{\mu}$ & $B_m$ & $B_h$ & $B_R$ & \\
\midrule
$0.002$ & 26.81\% & 72.22\% & 0.32\% & 0.65\% & 0.830 \\
$0.050$ & 22.54\% & 75.83\% & 0.54\% & 1.09\% & 0.602 \\
$0.100$ & 21.31\% & 76.47\% & 0.62\% & 1.60\% & 0.376 \\
$1.000$ & 29.35\% & 67.59\% & 0.86\% & 2.20\% & 0.206 \\
\bottomrule
\end{tabular}}
\end{table}

\textbf{Ablation Studies.} We conduct cumulative ablation experiments on the DL3DV dataset to evaluate the contribution of each component in GenSplatCodec, as summarized in Table~\ref{tab:ablation}. The Base model employs a DINO-based feed-forward Gaussian backbone to predict a renderable Gaussian representation, without the proposed Gaussian coding scheme or generative decoding approach. We then progressively introduce the components of detail-aware feed-forward Gaussian coding and geometry-guided one-step generative decoding.

We first examine the proposed Gaussian coding scheme. Adding the dual-branch Gaussian prediction improves the full-reference reconstruction metrics, increasing PSNR from 18.09 to 18.21 dB and reducing LPIPS from 0.411 to 0.407. Geometry-attribute decoupled coding then reduces the representation size from 15.784 to 4.679~MB, at the cost of a moderate fidelity degradation. By further introducing
rate-distortion-guided Gaussian compaction, the size is reduced to 0.577~MB while the reconstruction quality remains largely stable. These results demonstrate that the proposed coding scheme substantially reduces primitive redundancy and attribute coding cost while retaining
the principal scene structure.

We next evaluate the generative decoding components. Compared with the compressed Gaussian representation, the naive one-step generative decoder significantly improves perceptual quality, reducing LPIPS from 0.478 to 0.343 and FID from 111.84 to 79.32, but the decreased fidelity indicates that the generated details may not fully align with the transmitted scene structure. Dual-stream codec-aware adaptation subsequently improves both fidelity and perceptual quality by adapting the decoder to the structural and appearance cues recovered from the transmitted streams. Finally, hierarchical geometry control provides explicit structural constraints
for cross-view appearance transfer and achieves the best overall performance. Notably, the representation size remains 0.577~MB throughout the generative ablations, showing that these improvements introduce no additional transmission overhead.

Overall, compact Gaussian coding and geometry-guided generative decoding are complementary: the former preserves reliable scene structure under a tight rate budget, while the latter reconstructs perceptually faithful details from the decoded structural and appearance cues.

\textbf{Bit Allocation.}
Table~\ref{tab:quant}  reports the allocation of the total bitstream under different rate parameters on DL3DV. As the rate penalty increases, the average total size decreases from 0.830 to 0.206 MB, demonstrating flexible rate control. The main attribute stream $B_m$ accounts for 67.59\%-76.47\% of the total size and therefore constitutes the dominant coding cost, while the geometry stream $B_\mu$ accounts for 21.31\%-29.35\%. In contrast, the hyperprior and reference appearance streams together contribute no more than 3.06\% across all operating points. These results show that the coded reference appearance introduces only a small overhead, while most bits are allocated to the compact Gaussian structural representation.

\textbf{Runtime Analysis.} We evaluate the end-to-end codec runtime of GenSplatCodec. The average encoding and decoding times are 1.58~s and 1.80~s per scene, respectively. Encoding includes feed-forward Gaussian prediction and dual-stream coding, while decoding includes stream reconstruction, Gaussian rendering, and one-step generative decoding. The generative step takes only 0.12~s, avoiding iterative diffusion sampling. Moreover, independently coded Gaussian chunks enable multi-device parallelism while preserving intra-chunk context dependencies, allowing both encoding and decoding to be further accelerated beyond the reported single-device runtime.

\section{Conclusion}

This paper presented \textbf{GenSplatCodec}, a unified low-bitrate codec
that reformulates feed-forward Gaussian compression as geometry-guided
generative decoding. GenSplatCodec combines a compact Gaussian structural
stream produced by detail-aware feed-forward Gaussian coding with a
lightweight reference appearance stream. A geometry-guided one-step generative
decoding approach jointly exploits the decoded structural and appearance
cues to reconstruct perceptually faithful and view-consistent novel views without scene-specific optimization. The entire codec is optimized through a three-stage strategy for stable coding and generative adaptation. Experiments on DL3DV and RealEstate10K demonstrate superior rate-distortion performance and perceptual quality, substantially reduced storage costs, and strong view consistency.

\bibliographystyle{IEEEtran}
\bibliography{reference}

\end{document}